\documentclass[pdflatex,sn-mathphys-num]{sn-jnl}% Math and Physical Sciences Numbered Reference Style
\usepackage{graphicx}%
\usepackage{multirow}%
\usepackage{amsmath,amssymb,amsfonts}%
\usepackage{amsthm}%
\usepackage{mathrsfs}%
\usepackage[title]{appendix}%
\usepackage{xcolor}%
\usepackage{textcomp}%
\usepackage{manyfoot}%
\usepackage{booktabs}%
\usepackage{algorithm}%
\usepackage{algorithmicx}%
\usepackage{algpseudocode}%
\usepackage{listings}%
\theoremstyle{thmstyleone}%
\theoremstyle{thmstyletwo}%

\theoremstyle{thmstylethree}%

\begin{document}

\title[PERSEUS]{PERSEUS: Perception with Semantic Endoscopic Understanding and SLAM}
%Other option: LUMEN — Live Understanding and Mapping for Endoscopic Navigation


\author*[1]{\fnm{Ayberk} \sur{Acar}}\email{ayberk.acar@vanderbilt.edu}

\author[1]{\fnm{Fangjie} \sur{Li}}

\author[2]{\fnm{Susheela} \sur{Sharma Stern}}
\author[1]{\fnm{Lidia} \sur{Al-Zogbi}}
\author[1]{\fnm{Hao} \sur{Li}}
\author[1]{\fnm{Kanyifeechukwu Jane} \sur{Oguine}}
\author[1]{\fnm{Dilara}
\sur{Isik}}
\author[3]
{\fnm{Brendan}
\sur{Burkhart}}
\author[2]{\fnm{Jesse F.} \sur{d'Almeida}}
\author[2]{\fnm{Robert J.} \sur{Webster III}}
\author[1]{\fnm{Ipek} \sur{Oguz}}
\author[1]{\fnm{Jie Ying} \sur{Wu}}

\affil*[1]{\orgdiv{Department of Computer Science}, \orgname{Vanderbilt University}, \orgaddress{\street{2301 Vanderbilt Place}, \city{Nashville}, \postcode{37235}, \state{TN}, \country{USA}}}

\affil[2]{\orgdiv{Department of Mechanical Engineering}, \orgname{Vanderbilt University}, \orgaddress{\street{2301 Vanderbilt Place}, \city{Nashville}, \postcode{37235}, \state{TN}, \country{USA}}}

\affil[3]{\orgdiv{Laboratory for Computational Sensing and Robotics}, \orgname{Johns Hopkins University}, \orgaddress{\street{3400 N Charles St.}, \city{Baltimore}, \postcode{21218}, \state{MD}, \country{USA}}}


%\abstract{The abstract serves both as a general introduction to the topic and as a brief, non-technical summary of the main results and their implications. Authors are advised to check the author instructions for the journal they are submitting to for word limits and if structural elements like subheadings, citations, or equations are permitted.}

%%================================%%
%% Sample for structured abstract %%
%%================================%%

\abstract{\textbf{Purpose:} Natural orifice surgeries minimize the need for incisions and reduce the recovery time compared to open surgery; however, they require a higher level of expertise due to visualization and orientation challenges. We propose a perception pipeline for these surgeries that allows semantic scene understanding.

\textbf{Methods:} We bring learning-based segmentation, depth estimation, and 3D reconstruction modules together to create real-time segmented maps of the surgical scenes. Additionally, we use registration with robot poses to solve the scale ambiguity of mapping from monocular images, and allow the use of semantically informed real-time reconstructions in robotic surgeries.
 
\textbf{Results:} We achieve sub-milimeter reconstruction accuracy based on average one-sided Chamfer distances, average pose registration RMSE of 0.9 mm, and an estimated scale within 2\% of ground truth.
 
\textbf{Conclusion:} We present a modular perception pipeline, integrating semantic segmentation with real-time monocular SLAM for natural orifice surgeries. This pipeline offers a promising solution for scene understanding that can facilitate automation or surgeon guidance.}

\keywords{SLAM, 3D Reconstruction, Localization, Registration, Endoscopy, Segmentation, Depth Estimation, NOTES}



\maketitle

\section{Introduction}
\label{sec:intro}  % \label{} allows reference to this section
With the advancements in minimally invasive surgery technologies, new approaches such as endoluminal, transluminal, and natural orifice transluminal endoscopic surgeries (NOTES) have gained popularity~\cite{malik2006endoluminal}. NOTES aim to minimize the number of incisions and use natural access points in the body. NOTES allow operating through natural orifices such as the mouth, urethra, anus, or vagina, without additional incisions; however, at the cost of increasing complexity. Difficulties in visualization, spatial orientation, depth perception, and anatomy recognition make these procedures challenging, throttling expertise and practical applicability~\cite{swanstrom2008spatial}. These limitations show the significant need for better scene understanding to guide surgeons and automation. 

We present a perception pipeline for NOTES that incorporates real-time 3D reconstruction from monocular images, depth estimation, and segmentation. In our previous work~\cite{acar2025monocular}, we showed that offline monocular vision guidance can enable automated tumor removal. This work focuses on solving the limitations of the offline perception pipeline by creating a real-time solution to build segmented, scaled, and registered-to-surgical-scene 3D reconstructions of a surgical field. Furthermore, we extend the evaluation of 3D reconstruction to a different anatomy to demonstrate generalizability and comparatively evaluate mapping methods. Codebase for this project is available on GitHub: \url{github.com/vu-maple-lab/perseus}

\section{Background and Related Work}
\textbf{Clinical Relevance: }
NOTES technique, originally developed for abdominal surgeries, later gained interest in different application areas such as thoracic or urologic surgeries~\cite{makris2010natural, tyson2014urological}. In this study, we focus on two clinical problems: central airway obstruction (CAO), and benign prostatic hyperplasia (BPH), which can be treated via oral and urethral access, respectively.

CAO is a disorder with increasing prevalence and is a cause of significant morbidity and mortality~\cite{ernst2004central}. Therapeutic tools such as laser and electrocautery that are delivered through bronchoscopes can enable complete removal of the obstructing tumor~\cite{ernst2004central}, but the procedure is challenging and complications can be fatal~\cite{stahl2015complications, gafford2020concentric}.

BPH refers to growth of prostate lobes and is very prevalent in aging males~\cite{roehrborn2005benign}. Lower urinary tract symptoms and bladder outlet obstruction (BOO) can stem from BPH, which can affect patients' quality of life. BOO may interfere with sexual functioning and potentially cause urinary tract infections, formation of bladder stones, and renal failures~\cite{roehrborn2005benign}. Treatment options for BPH vary from medical therapy to surgical interventions such as transurethral resection of prostate.%~\cite{lokeshwar2019epidemiology}.

Recent studies show that robotic systems can be used for CAO removal with reduced complications~\cite{gafford2020concentric}. Additionally, with proper visual guidance, automation of this procedure is feasible, allowing precise and consistent performance. Previous studies show the feasibility of tumor removal on a benchtop setting with open surgery models~\cite{acar2025monocular,smith2025autonomous}. Autonomous features for robotic BPH treatment techniques, such as using a robotic arm to direct a high-pressure water jet for tissue removal, may improve outcomes~\cite{connor2020autonomous}. Accurate mapping and segmentation for transurethral approaches can extend automation to other surgeries.

\textbf{3D Reconstructions: }
In our previous work, we combined 3D reconstruction of monocular images with segmentation methods to guide the automation of tumor removal~\cite{acar2025monocular}. However, the Structure-from-Motion (SfM) method used in our previous study is time-consuming, and repeating the reconstructions in case of failure, or as the operation progresses, is not feasible. 

Simultaneous Localization and Mapping (SLAM) algorithms offer a good alternative to SfM, allowing real-time reconstructions and camera pose estimations. Previous studies show the feasibility of SLAM applications in a wide range of laparoscopic and endoscopic surgeries such as sinus endoscopy, colonoscopy, and ureteroscopy~\cite{teufel2024oneslam, liu2022sage, oliva2025kidneydepth}. Previous work demonstrated that integration of explicit depth estimation and priors can improve SLAM reconstruction results compared to using monocular video alone~\cite{liu2022sage, oliva2025kidneydepth}. In terms of segmentation integration into SLAM, literature mostly focused on outlier detection, use of semantic information to improve SLAM performance, or creating segmented scenes for natural images~\cite{liu2021rds, tateno2015real}. Other work proposed use of segmentation to remove dynamic tools in the surgical scenario, to reduce reconstruction errors~\cite{wu2022semantic}. However, literature focusing on use of semantic information to detect target anatomies and create informed surgical scenes is very limited. Methods such as Semantic-SuPer bring together information from depth estimation, semantic segmentation, and tool poses to create segmented surgical scenes~\cite{lin2023semantic}. This method is tested only on an open access laparoscopy setup and not implemented for endoscopy. 

In this study, we combine the real-time 3D reconstructions with semantic target information acquired from a deep learning model. This allows fast and accurate scene understanding in endoscopic surgeries, enabling downstream tasks such as surgical automation. To the best of our knowledge, this study is the first to use real-time SLAM methods for CAO and BPH, creating new possibilities for surgical approaches. 

\begin{figure}[tbp]
\centering
\includegraphics[width=0.8\columnwidth]{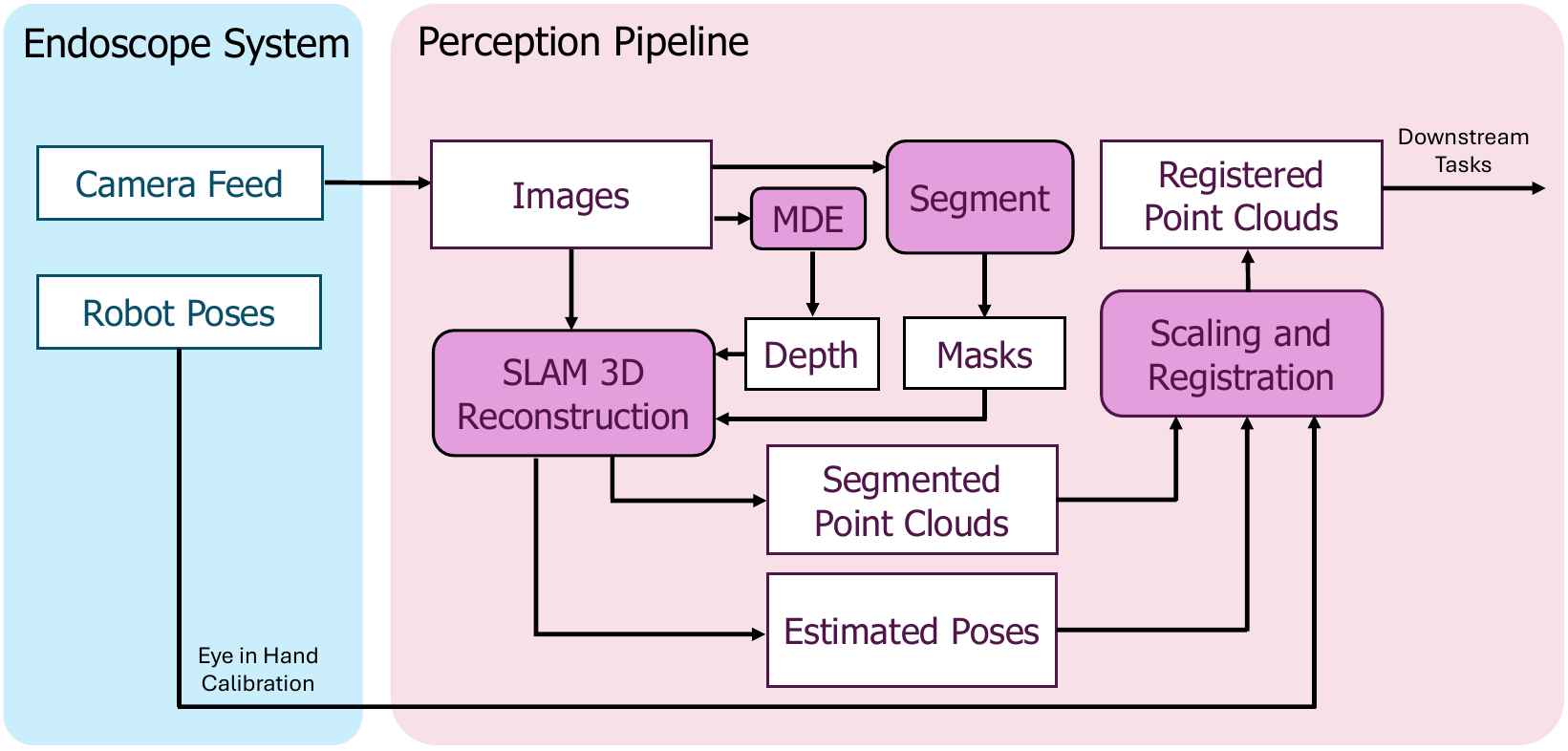}
\caption{Diagram showing the overall workflow. We use depth estimations and target segmentation masks in the SLAM system to create segmented point clouds of the surgical scene. We use estimated and robot camera poses to scale and register these reconstructions, to be used in downstream tasks.}
\label{fig:overall_workflow}
\end{figure}
%One combo figure showing image, segmented image, reconstruction, segmented reconstruction

\section{Methods}
\label{sec:methods}

An overall workflow can be seen in Figure~\ref{fig:overall_workflow}. This section details the modules of our workflow and the data collection. We explain the reconstruction algorithm in detail in subsection~\ref{sec:reconstruction}, segmentation and MDE model in~\ref{sec:seg_mde}, segmented point cloud generation in~\ref{sec:segmented_recons} and scaling and registration in~\ref{sec:registration}.

\subsection{Phantom Production and Data Collection}
\begin{figure}[tbp]
\centering
\includegraphics[width=0.5\columnwidth]{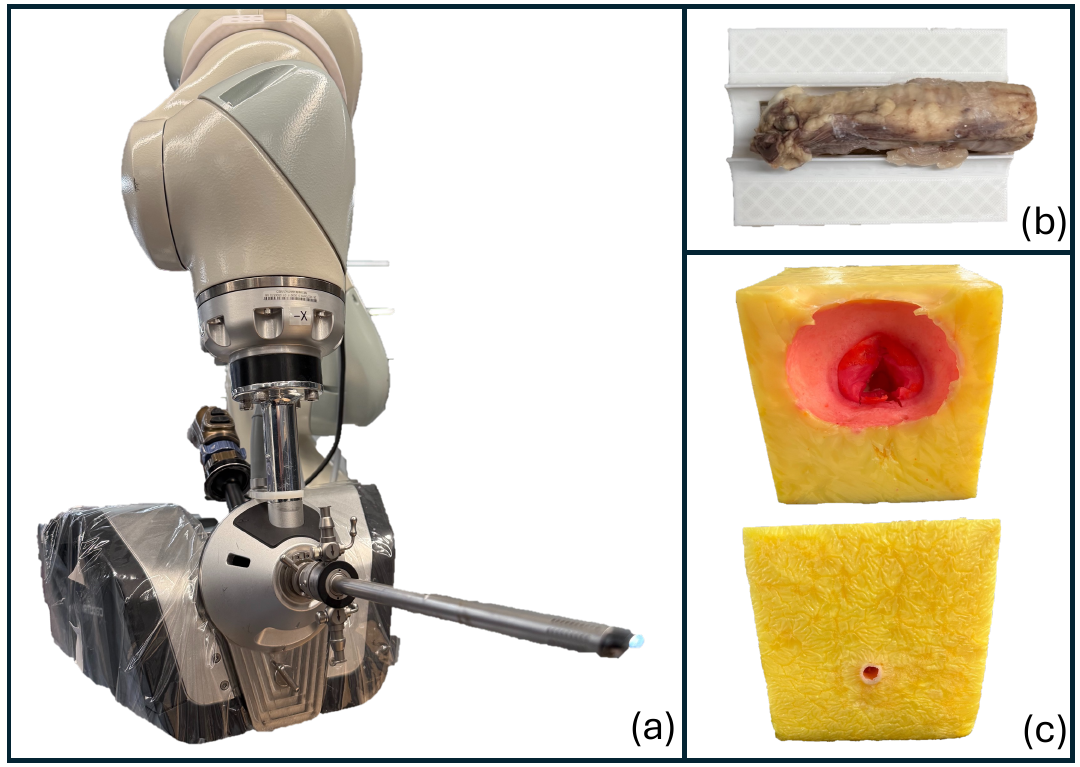}
\caption{Camera system and phantoms used. Endoscope held by robotic arm \textbf{(a)} is inserted to trachea phantom from the end points \textbf{(b)} and prostate phantom from the urethra \textbf{(c)} to collect videos.}
\label{fig:setup}
\end{figure}

For the CAO case, following our previous work~\cite{acar2025monocular}, we prepare phantoms using sheep pluck and chicken breast to mimic human anatomy. We separate the trachea from the rest of the pluck for easier handling. We cut small pieces of chicken breast that cause around 50\% occlusion in the trachea, place them inside the airway through small incisions, and secure them with super glue (Fig.~\ref{fig:setup}b). 

For our BPH experiments, we use hydrogel phantoms developed originally for surgical training~\cite{saba2022design}. These phantoms are modeled after patient CT scans, and reflect the visual and material properties of human anatomy (Fig.~\ref{fig:setup}c). 

We record monocular videos on both phantoms with Virtuoso Surgical (Nashville, TN, USA) endoscopy system (Fig.~\ref{fig:setup}a). Additionally, to acquire ground truth geometries for evaluation of our perception pipeline, we take computed tomography (CT) scans of the phantoms and segment the anatomies manually. 

\subsection{Segmentation and Monocular Depth Estimation}
\label{sec:seg_mde}
We train two separate segmentation models for CAO tumor identification and BPH lobe differentiation. Although the models share an identical network architecture and training configuration, they differ in the dataset composition used for training.

Our segmentation framework builds upon a U-Net–style~\cite{ronneberger2015u} architecture integrated with SAM2~\cite{ravi2024sam} as the encoder backbone. %This design leverages SAM2’s strong feature representation and generalization capabilities to maintain robust performance under complex surgical conditions, such as occlusions, lighting variations, and motion artifacts. 
To efficiently adapt SAM2 to the CAO domain, we employ adapter-based fine-tuning~\cite{li2024promise}, where lightweight adapter modules are inserted into the transformer blocks to update task-specific parameters while keeping the majority of the pretrained weights frozen. The fine-tuning of SAM2 follows the approach outlined in our recent work~\cite{jane2026segmentation}, which provides comprehensive details on the model design, data preprocessing, objective function, and optimization strategy for CAO segmentation. The same architectural design and training strategy were applied to the BPH segmentation task. In this case, the model was trained on approximately 3,300 images, containing a diverse range of surgical actions on the phantoms, from exploratory views of the BPH anatomy to operative environments involving tool interactions and tissue deformation. %This diversity in training data ensures that the model generalizes effectively across different stages and visual complexities of the BPH procedure.

To address the lack of ground truth depth in endoscopic images, we use an MDE framework~\cite{li2025monocular} that integrates 3D CT data and 2D endoscopic video data. Our MDE consists of three main stages: (1) Rendered data generation in Unity using 3D CT segmentations of CAO and BPH scans to produce absolute depth maps~\cite{lu2025kidney}; (2) Unsupervised domain adaptation~\cite{li2024deep} between rendered and real endoscopy data, where rendered images are translated into the visual style of real endoscopic scenes to reduce the domain gap; and (3) Depth prediction by fine-tuning the DepthAnythingV2 model~\cite{yang2024depth} with paired translated rendered images and depth maps from step (2), achieving accurate depth estimation. Separate models were trained for CAO and BPH, respectively. Further analysis for segmentation~\cite{jane2026segmentation} and MDE~\cite{li2025monocular} models can be found in corresponding publications.

\begin{figure}[tbp]
\centering
\includegraphics[width=0.8\columnwidth]{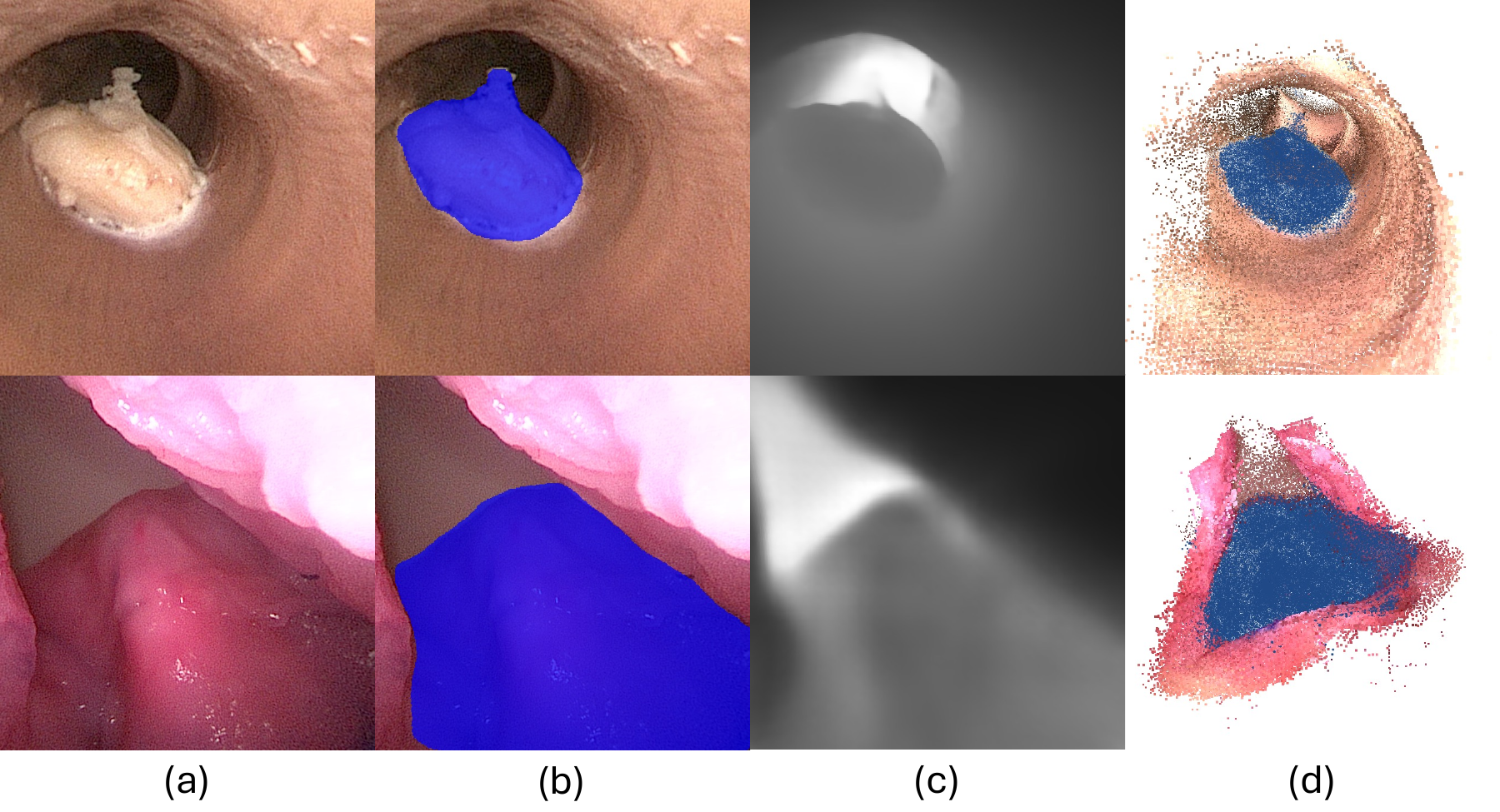}
\caption{\textbf{(a)} Example endoscope images \textbf{(b)} Segmentation overlays  \textbf{(c)} Monocular depth estimation results \textbf{(d)} Segmented 3D reconstructions. Top row shows central airway obstruction and bottom row shows the benign prostatic hyperplasia case.}
\label{fig:image_and_reconstruction}
\end{figure}
%One combo figure showing image, segmented image, reconstruction, segmented reconstruction

\subsection{Real time 3D Reconstruction}
\label{sec:reconstruction}
To create 3D reconstructions in real time, we use DROID-SLAM~\cite{teed2021droid}. DROID-SLAM uses deep learning to iteratively update and correct estimated camera poses and predicted per-pixel depth values for the corresponding frames. The corrected depth values are later mapped to a point cloud in 3D using an inverse projection function and the pinhole camera model. DROID-SLAM supports multiple input modes, such as RGB-D or stereo. To augment our monocular video, we integrate our customized in-domain MDE model to create priors for the SLAM algorithm. This results in a pseudo-RGB-D pipeline (referred to as RGB-D SLAM), without the bulky and expensive depth cameras. We use the DROID-SLAM model weights from~\cite{teed2021droid} without additional training, testing the generalization capability of the reconstruction algorithm.

We perform a checkerboard calibration before explorations to obtain the camera parameters required by the SLAM framework. Since the endoscope has a fisheye effect for an increased field of view, following the original DROID-SLAM implementation, we undistort the camera stream. We crop the images around the image center to remove the black mask around the endoscope view. Then, we scale these preprocessed images to match SLAM's model input size. During the reconstruction, for the inverse projection, we use the estimated rectified camera matrix acquired from the camera calibration for the corresponding undistorted images. 

Using DROID-SLAM with the monocular image stream from the endoscope allows us to create point clouds in real-time. However, these reconstructions do not have the crucial context information, such as tumor borders, required to identify the regions of interest for automation. We integrate a segmentation pipeline to the reconstruction process, to create semantically informed scene maps.

\subsection{Segmented Reconstructions}
\label{sec:segmented_recons}
To identify the regions associated with the targeted areas in the point cloud, we run a segmentation model inference in each frame in the image stream. At the inverse projection step of the reconstruction algorithm, we rescale the segmentation masks to match the size of the final inverse depth maps and use them to index the 3D point cloud acquired from depth images. Points corresponding to the pixels fall into the segmentation region are stored separately, and visualized in a different color (Fig.~\ref{fig:image_and_reconstruction}d).

\subsection{Registration and Scaling the Reconstruction}
\label{sec:registration}
Although we use an MDE model that outputs absolute depth estimation, imperfections in this model can still cause scale differences in the monocular reconstruction. While these reconstructions are still useful as a 3D map, correct scale is required for applications such as automation and measurements. To overcome the scale ambiguity and register the estimated map onto the surgical scene, we use robot kinematics data of the camera motion. By reading the robot arm poses and applying an eye-in-hand calibration~\cite{park1994robot}, we acquire the robot-held-camera pose readings. We use the Umeyama~\cite{umeyama1991least} algorithm to register the translation components of estimated camera poses from SLAM (post bundle adjustment) to robot-held-camera pose readings. Registration of these two independently-acquired poses avoids the fiducials required in our previous work~\cite{acar2025monocular} and allows immediate scaling and alignment of the segmented 3D maps to the physical scene. This method can be generalized to non-robotic applications by acquiring the camera poses using means such as optical trackers.

\begin{figure}[bp]
\centering
\includegraphics[width=0.9\columnwidth]{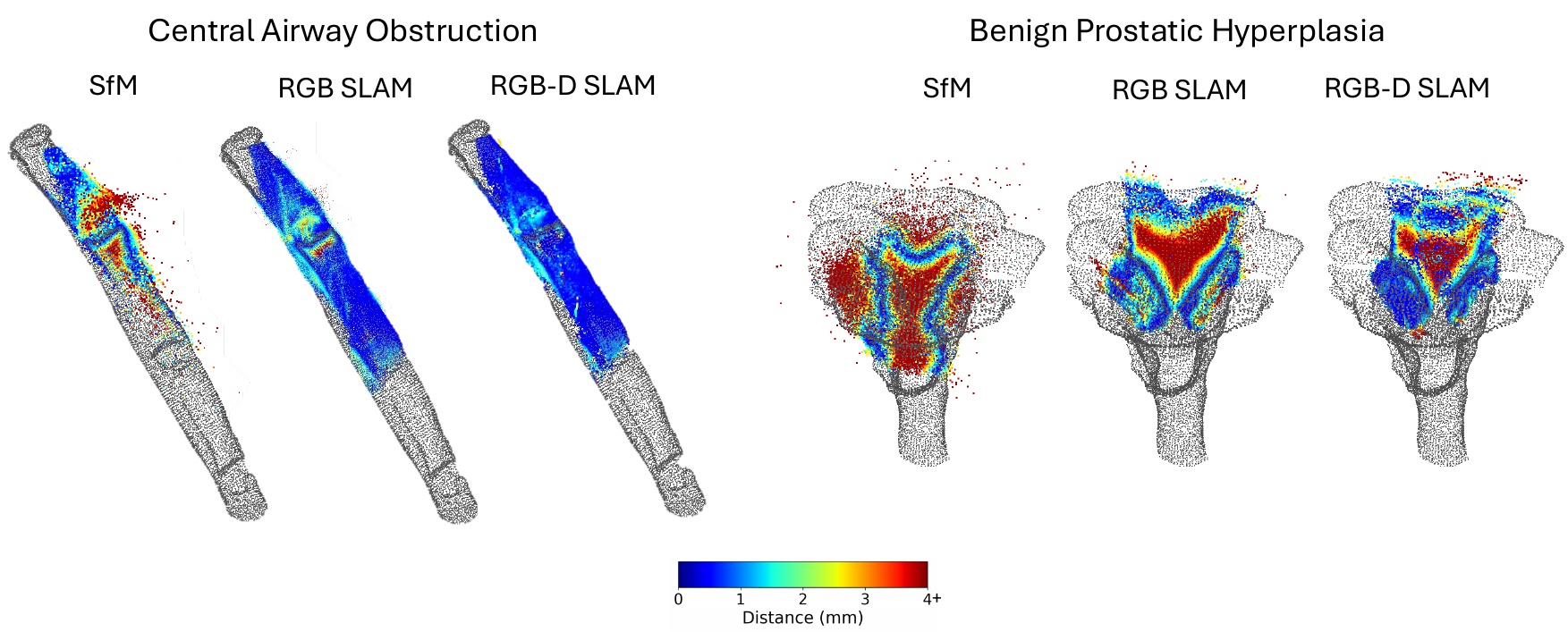}
\caption{Example evaluation of reconstructions with registration to CT scan point cloud. Heatmaps indicate distance to the closest point.}
\label{fig:ClosestPoint}
\end{figure}
\section{Experiments and Results}
\label{section:results}
%\textcolor{red}{ADD: TWO TABLES.
%TABLE1: CAO: SFM, WITH OR WITHOUT MDE SLAM, SAME METRICS. SFM NON-FAILED ONES BUT ALSO WRITE HOW MANY FAILED. TABLE2: SAME STUFF WITH BPH.}

\begin{table}[]
\caption{Average quantitative evaluation results for 3D reconstruction methods. $\uparrow$ indicates higher is better, $\downarrow$ indicates lower is better. Best results in the category are given bold and the second best is underlined. SLAM processing times are reported before and after bundle adjustment (BA).}
\label{table:quantresults}

\begin{tabular}{|c|c|c|c|}
\hline
\textbf{Central Airway Obstruction}           & \textbf{SfM}  & \textbf{RGB SLAM}         & \textbf{RGB-D SLAM}       \\ \hline
Median Closest Point Dist. (mm) $\downarrow$  & 0.63 $\pm$ 0.13  & \textbf{0.52 $\pm$ 0.23}              & \underline{0.53 $\pm$ 0.32}            \\ \hline
One-sided Chamfer Dist. (mm) $\downarrow$      & 0.90 $\pm$ 0.17  & \underline{0.67 $\pm$ 0.30}             & \textbf{0.66 $\pm$ 0.40}             \\ \hline
One-Sided Hausdorff Dist.  (mm) $\downarrow$   & 100.96 $\pm$ 127.29  & \underline{15.10 $\pm$ 5.27}              & \textbf{6.92 $\pm$ 2.24}              \\ \hline
Pre-BA Time Per Frame (sec) $\downarrow$& - & \textbf{0.26 $\pm$ 0.02}  & \underline{0.29 $\pm$ 0.02} \\ \hline
Post-BA Time Per Frame (sec) $\downarrow$ & 6.39 $\pm$ 1.98 & \textbf{0.42 $\pm$ 0.07} & \underline{0.46 $\pm$ 0.07} \\ \hline
Segmentation Precision (\%) $\uparrow$          & \textbf{96.69 $\pm$ 1.59}  & 85.48 $\pm$ 5.09              & \underline{88.07 $\pm$ 2.56}             \\ \hline
\# Reconstruct. Points (x1000)  $\uparrow$        & 121.6 $\pm$ 44.6   & \textbf{1021.2 $\pm$ 259.3}               & \underline{588.1 $\pm$ 130.0}               \\ \hline
 Coverage (\%)   $\uparrow$    & 27.76 $\pm$ 11.97   & \textbf{43.84 $\pm$ 5.97}              & \underline{35.55 $\pm$ 6.14}              \\ \hline
\textbf{Benign Prostatic Hyperplasia}                 & \textbf{SfM}  & \textbf{RGB SLAM}         & \textbf{RGB-D SLAM}       \\ \hline
Median Closest Point Dist. (mm) $\downarrow$   & 1.86 $\pm$ 0.30  & \underline{0.63 $\pm$ 0.19}              & \textbf{0.52 $\pm$ 0.15}             \\ \hline
One-sided Chamfer Dist. (mm) $\downarrow$      & 2.25 $\pm$ 0.40  & \underline{0.89 $\pm$ 0.19}             & \textbf{0.65 $\pm$ 0.19}             \\ \hline
One-Sided Hausdorff Dist. (mm)  $\downarrow$   & 31.63 $\pm$ 13.95  & \underline{21.25 $\pm$ 14.61}              & \textbf{9.95 $\pm$ 1.78}              \\ \hline
Pre-BA Time Per Frame (sec) $\downarrow$ & -  & \textbf{0.21 $\pm$ 0.05}   & \underline{0.23 $\pm$ 0.05}  \\ \hline
Post-BA Time Per Frame (sec) $\downarrow$ & 16.76 $\pm$ 8.70 & \textbf{0.26 $\pm$ 0.05}  & \underline{0.28 $\pm$ 0.05} \\ \hline
Segmentation Precision (\%) $\uparrow$            & \textbf{96.23 $\pm$ 2.64}  & 89.71 $\pm$ 5.18              & \underline{90.87 $\pm$ 5.40}              \\ \hline
\# Reconstruct. Points (x1000)  $\uparrow$         & 84.3 $\pm$ 50.7   & \textbf{467.5 $\pm$ 236.3}               & \underline{288.3 $\pm$ 134.6}               \\ \hline
Coverage (\%)     $\uparrow$                    & \underline{20.85 $\pm$ 2.53}  & \textbf{23.52 $\pm$ 4.21}              & 20.13 $\pm$ 2.65              \\ \hline
\end{tabular}
\end{table}

\subsection{Reconstruction Quality}

We comparatively and quantitatively evaluate the reconstruction accuracy and the segmentation precision on five exploration videos acquired from three different CAO phantoms and two exploration videos acquired from two different BPH phantoms. Extracted frame numbers vary from 512 to 1114. To quantify the reconstruction accuracy independently from other components, we scale and register the 3D reconstructions to the segmented ground-truth CT point cloud manually using visual cues. We initialize the position and scale of the reconstructions and fine-tune the registration with Iterative Closest Point (ICP) algorithm (Fig.~\ref{fig:ClosestPoint}).

Our baseline for comparison is the DISK~\cite{tyszkiewicz2020disk}+LightGlue~\cite{lindenberger2023lightglue} SfM pipeline of the Hierarchical Localization Toolkit~\cite{sarlin2019coarse}, used in our previous work~\cite{acar2025monocular}. This time, for a fair comparison, we also undistorted the images and center-cropped the endoscope mask same way as the SLAM pipeline input before feeding into the SfM algorithm. We also compare the performance of SLAM with and without the MDE integration.

Table~\ref{table:quantresults} shows the quantitative results. SLAM algorithm results are post global bundle adjustment. Coverage means the percentage of CT point cloud points that have a correspondence in reconstruction, within 1\,mm distance. The one-sided Chamfer and Hausdorff distances are calculated from reconstruction to CT scan models. Figure~\ref{fig:ClosestPoint} shows qualitative evaluations of one sequence for each phantom case, for the closest point metric. Experiments are done on an NVIDIA RTX 4090 GPU. During our real-time experiments, we do not observe a significant delay before the global bundle adjustment step that takes place at the end of the image stream.

\begin{figure}[tbp]
\centering
\includegraphics[width=0.95\columnwidth]{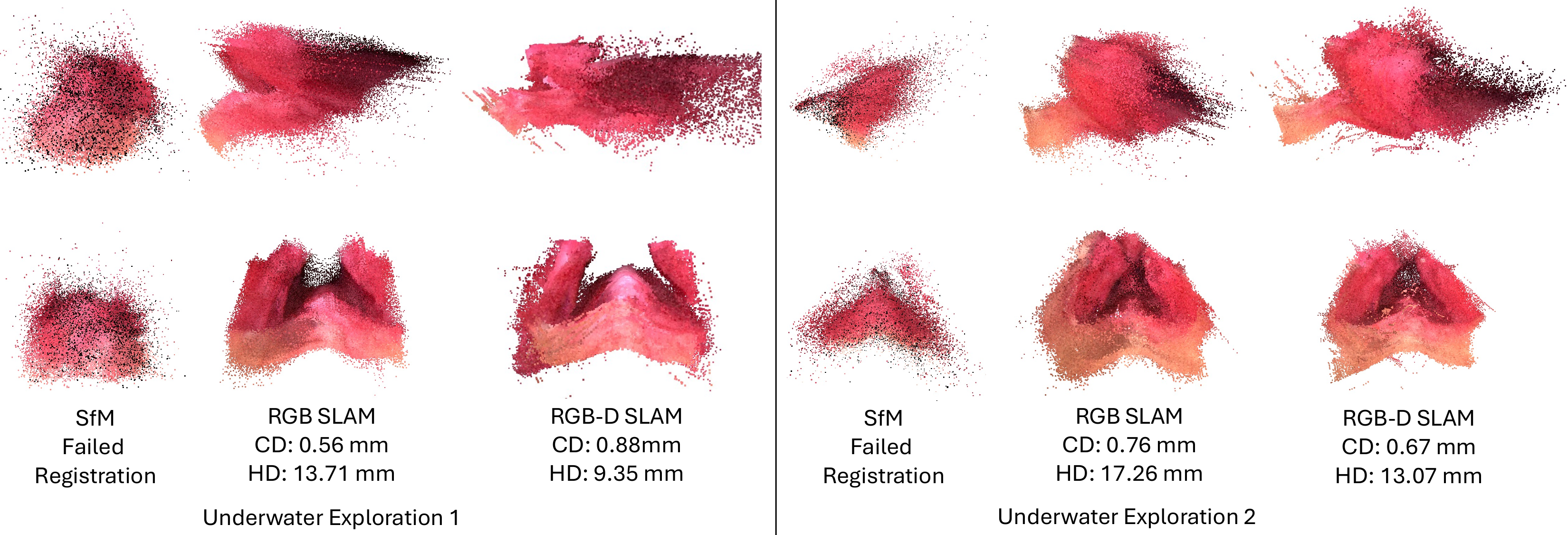}
\caption{Reconstruction quality evaluation on submerged prostate phantoms. CD: One-Sided Chamfer Distance, HD: One-Sided Hausdorff Distance.}
\label{fig:UnderWater}
\end{figure}

To calculate the segmentation precision, we project the segmented tumor point cloud, that contains points from all of the images, back to the segmentation masks, using the estimated poses from reconstruction algorithm and the pinhole camera model. We analyze the successfully registered frames for SfM, and keyframes for SLAM. The ratio of projected points that fall within the segmentation masks over the total gives us the precision of the segmentation projection. This reprojection process allows us to analyze the combined segmented points from different frames, and the effect of reconstruction algorithm through bundle adjustments, independently from the success of segmentation model. We note that this evaluation method can be affected by projection of points associated with the areas occluded in the projected image, however, we evaluate all the methods consistently.

By changing the reconstruction parameters affecting the number of keyframes, point selection thresholds, and the number of features in the SfM pipeline, the balance between reconstruction speed and quality, or reconstruction density and noise can be changed. Finally, we note that coverage depends on the extent of exploration in the video and is only useful as a comparison between the methods.

\begin{figure}[tbp]
\centering
\includegraphics[width=0.9\columnwidth]{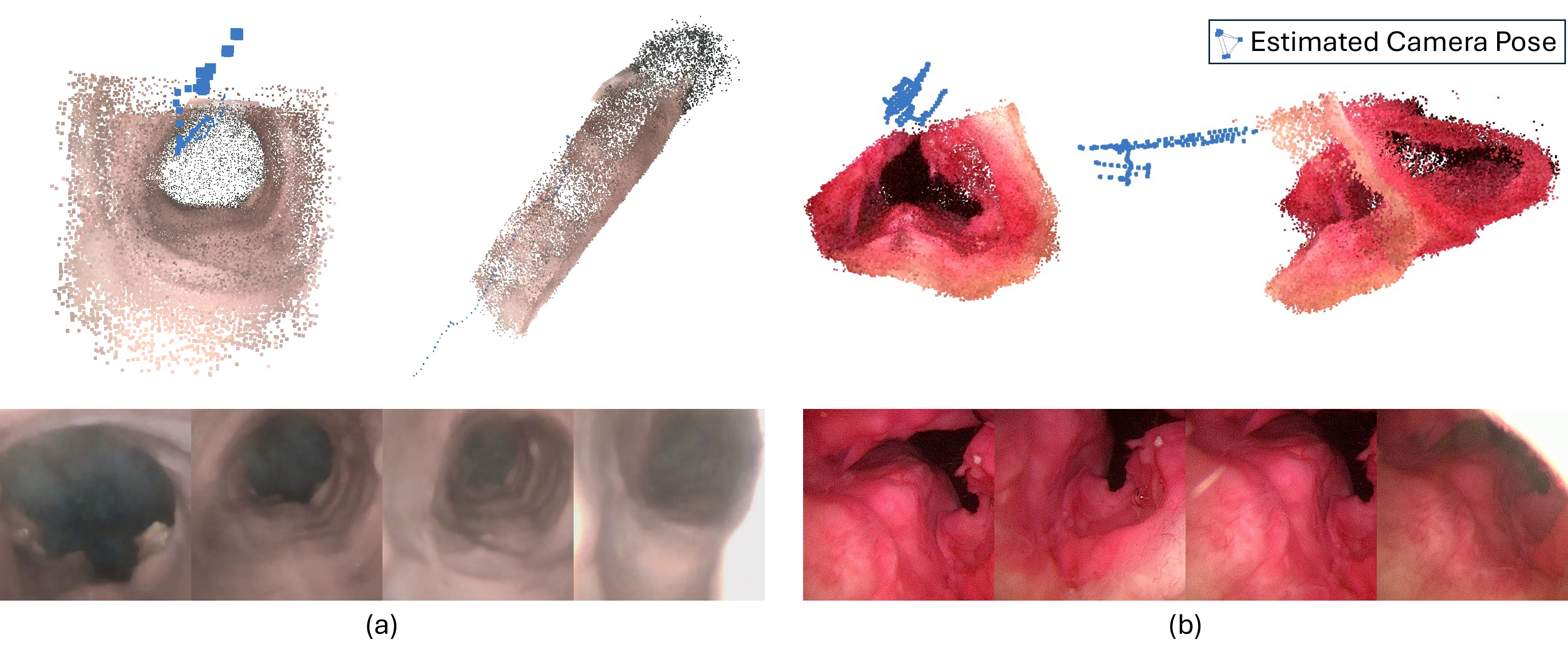}
\caption{Qualitative examples showing reconstruction generalizability with RGB-D SLAM reconstructions and example frames from corresponding videos. \textbf{(a)} Reconstruction of a cadaver central airway. \textbf{(b)} Reconstruction of submerged prostate phantom, mid-operation, after one of the lobes is resected.}
\label{fig:QualResults}
\end{figure}

Besides the comparative analysis in Table~\ref{table:quantresults}, we collected one operation and two exploration videos for the case when a prostate phantom is submerged underwater, and one cadaver central airway exploration video. This data is excluded from the comparative analysis due to a lack of registration visual cues in SfM reconstructions. Prostate exploration reconstruction analysis is given in Figure~\ref{fig:UnderWater}, and qualitative results for cadaver and mid-operation reconstruction are given in Figure~\ref{fig:QualResults}.

\subsection{Registration Accuracy}
% \usepackage{graphicx}

For registration and scaling evaluation, we collect one more exploration sequence with robot camera poses for each anatomy. To evaluate the accuracy of the registration, we calculate the root mean square error (RMSE) between robot-held camera poses and estimated registered camera poses from RGB-D SLAM algorithm. This residual error shows how well SLAM reconstructed the camera trajectory. To measure the accuracy of scaling, we cauterize small fiducial points in the anatomy, and compare the real distances measured by calipers vs. estimated distances for 5 point pairs. Average of estimated distances over measured distances give us the scale ratio, which would be 100\% in the ideal scenario. We estimate a global scale factor for the reconstruction which does not affect the reconstruction quality. We note that accuracy of scaling and registration can be affected by the kinematics accuracy of the industrial arm holding the endoscope, and calibration quality.

\begin{table}[b]
\centering
\caption{Evaluation of registration and scaling accuracy.}
\begin{tabular}{|c|c|c|}
\hline
               & \textbf{Scale Ratio (\%)} & \textbf{Pose RMSE (mm)} \\ \hline
Central Airway &         98.17   $\pm$    3.49        &          0.67         \\ \hline
Prostate       &        98.23    $\pm$    5.88      &           1.12         \\ \hline
\end{tabular}%

\end{table}

\section{Discussion}
From the quantitative results in Table~\ref{table:quantresults}, we see that the use of the SLAM algorithm results in quicker and more accurate reconstructions than the baseline SfM pipeline. The dense reconstruction nature of the SLAM algorithm, which relies on depth estimation, helps extend coverage to areas with a smaller number of observations and increases the number of reconstructed points. While both SLAM methods have higher reconstruction density and coverage compared to SfM algorithm, a decrease in average number of reconstruction points and coverage in the RGB-D SLAM compared to the RGB SLAM can be attributed to lower noise in the RGB-D reconstruction. This lower noise level can be understood better using the Hausdorff distance metric. However, real-time projection of segmentation masks without additional modification, such as weighting different masks as in~\cite{acar2025monocular}, causes a lower segmentation precision. In both the CAO and the BPH cases, error metrics show similar trends.

The performance of this pipeline is heavily dependent on its learning-based components. For the underwater scenario (Fig.~\ref{fig:UnderWater}), the depth model decreases the noise, leading to a smaller Hausdorff distance. However, since the underwater views are affected by the domain gap between images in air and water, the use of MDE results in larger Chamfer Distance and does not necessarily improve the reconstructions. Nonetheless, the overall pipeline can be easily adapted to different scenarios or procedures, and shows a promising performance in cadaver experiment without any modifications (Fig.~\ref{fig:QualResults}a). Although DROID-SLAM shows a good generalization for the endoscopic cases, with in-domain training of networks in the SLAM pipeline and better depth estimation models, the reconstruction qualities could be improved.

While SfM, with its offline nature, is more robust to temporary occlusions or other factors that can cause tracking to reinitialize, recovery in real-time for SLAM is more challenging. This can also affect the segmentation precision and reconstruction accuracy. Even though the SLAM algorithm is able to reconstruct in the scenarios SfM failed (Fig.~\ref{fig:UnderWater}), modifications to real-time reconstruction pipeline may be needed in the clinical scenario where factors such as blood or debris are more apparent.

Finally, as a limitation, this pipeline does not take the scene deformations into account and as the operation progresses, surgical scenes change significantly. Therefore, intermediate reconstructions can be used to update the scene representations (e.g. in Fig.~\ref{fig:QualResults}b). This limitation is less apparent in CAO due to the rigid nature of the trachea and as can be seen in our previous work~\cite{acar2025monocular}, surgical automation is still possible. Our future work includes improving the perception with deformation modeling and integration of other modalities, such as ultrasound. We plan to test the perception pipeline in downstream tasks and clinical videos.

\section{Conclusion}
\label{sec:conclusion}
In this study, we presented a pipeline to create segmented 3D reconstructions for NOTES. These 3D maps can aid scene understanding and automation. We evaluated our pipeline by comparing with the ground truth geometries obtained from CT scans and observed improvement in both processing times and reconstruction accuracies. Due to the plug-and-play nature of the pipeline and generalization capabilities of the used SLAM architecture, the methods presented in this paper can be extended to other anatomies or surgeries by simply changing the segmentation and depth estimation models, also making the overall proposed pipeline agnostic to domain shifts. 

\section*{Declarations}
\begin{itemize}
\item Authors thank Nithesh Kumar and Alejandro Chara for their contributions to data collection, and Lauren Branscombe and Josh Petrin for their technical support.
\item Funding: Research reported in this publication was supported by the Advanced Research Projects Agency for Health (ARPA-H) under Award Number D24AC00415-00. The ARPA-H award provided 100\% of total costs with an award total of up to \$11,935,038. The content is solely the responsibility of the authors and does not necessarily represent the official views of ARPA-H.
\item Competing Interests: No relevant financial or non-financial interests to disclose.
\end{itemize}

\bibliography{sn-bibliography}% common bib file

@article{ernst2004central,
  title={Central airway obstruction},
  author={Ernst, Armin and Feller-Kopman, David and Becker, Heinrich D and Mehta, Atul C},
  journal={American journal of respiratory and critical care medicine},
  volume={169},
  number={12},
  pages={1278--1297},
  year={2004},
  publisher={American Thoracic Society}
}

@article{stahl2015complications,
  title={Complications of bronchoscopy: A concise synopsis},
  author={Stahl, David L and Richard, Kathleen M and Papadimos, Thomas J},
  journal={International journal of critical illness and injury science},
  volume={5},
  number={3},
  pages={189--195},
  year={2015},
  publisher={Medknow}
}

@inproceedings{li2024promise,
  title={Promise: Prompt-driven 3d medical image segmentation using pretrained image foundation models},
  author={Li, Hao and Liu, Han and Hu, Dewei and Wang, Jiacheng and Oguz, Ipek},
  booktitle={2024 IEEE international symposium on biomedical imaging (ISBI)},
  pages={1--5},
  year={2024},
  organization={IEEE}
}

@article{ravi2024sam,
  title={Sam 2: Segment anything in images and videos},
  author={Ravi, Nikhila and Gabeur, Valentin and Hu, Yuan-Ting and Hu, Ronghang and Ryali, Chaitanya and Ma, Tengyu and Khedr, Haitham and R{\"a}dle, Roman and Rolland, Chloe and Gustafson, Laura and others},
  journal={arXiv preprint arXiv:2408.00714},
  year={2024}
}

@article{gafford2020concentric,
  title={A concentric tube robot system for rigid bronchoscopy: a feasibility study on central airway obstruction removal},
  author={Gafford, Joshua B and Webster, Scott and Dillon, Neal and Blum, Evan and Hendrick, Richard and Maldonado, Fabien and Gillaspie, Erin A and Rickman, Otis B and Herrell, S Duke and Webster III, Robert J},
  journal={Annals of biomedical engineering},
  volume={48},
  number={1},
  pages={181--191},
  year={2020},
  publisher={Springer}
}

@article{smith2025autonomous,
  title={Autonomous vision-guided resection of central airway obstruction},
  author={Smith, Mariana E and Yilmaz, Nural and Watts, Tanner and Scheikl, Paul Maria and Ge, Jiawei and Deguet, Anton and Kuntz, Alan and Krieger, Axel},
  journal={arXiv preprint arXiv:2502.18586},
  year={2025}
}

@inproceedings{acar2025monocular,
  title={From monocular vision to autonomous action: Guiding tumor resection via 3d reconstruction},
  author={Acar, Ayberk and Smith, Mariana and Al-Zogbi, Lidia and Watts, Tanner and Li, Fangjie and Li, Hao and Yilmaz, Nural and Scheikl, Paul Maria and d’Almeida, Jesse F and Sharma, Susheela and others},
  booktitle={2025 IEEE/RSJ International Conference on Intelligent Robots and Systems (IROS)},
  pages={21714--21720},
  year={2025},
  organization={IEEE}
}

@article{teed2021droid,
  title={Droid-slam: Deep visual slam for monocular, stereo, and rgb-d cameras},
  author={Teed, Zachary and Deng, Jia},
  journal={Advances in neural information processing systems},
  volume={34},
  pages={16558--16569},
  year={2021}
}

@article{malik2006endoluminal,
  title={Endoluminal and transluminal surgery: current status and future possibilities},
  author={Malik, A and Mellinger, JD and Hazey, JW and Dunkin, BJ and MacFadyen Jr, BV},
  journal={Surgical Endoscopy and Other Interventional Techniques},
  volume={20},
  number={8},
  pages={1179--1192},
  year={2006},
  publisher={Springer}
}

@inproceedings{makris2010natural,
  title={Natural orifice trans-luminal endoscopic surgery (NOTES) in thoracic surgery},
  author={Makris, Konstantinos I and Rieder, Erwin and Swanstrom, Lee L},
  booktitle={Seminars in thoracic and cardiovascular surgery},
  volume={22},
  number={4},
  pages={302--309},
  year={2010},
  organization={Elsevier}
}

@article{tyson2014urological,
  title={Urological applications of natural orifice transluminal endoscopic surgery},
  author={Tyson, Mark D and Humphreys, Mitchell R},
  journal={Nature Reviews Urology},
  volume={11},
  number={6},
  pages={324--332},
  year={2014},
  publisher={Nature Publishing Group UK London}
}

@article{swanstrom2008spatial,
  title={Spatial orientation and off-axis challenges for NOTES},
  author={Swanstrom, Lee and Zheng, Bin},
  journal={Gastrointestinal endoscopy clinics of North America},
  volume={18},
  number={2},
  pages={315--324},
  year={2008},
  publisher={Elsevier}
}

@article{roehrborn2005benign,
  title={Benign prostatic hyperplasia: an overview},
  author={Roehrborn, Claus G},
  journal={Reviews in urology},
  volume={7},
  number={Suppl 9},
  pages={S3},
  year={2005}
}

@article{connor2020autonomous,
  title={Autonomous surgery in the era of robotic urology: friend or foe of the future surgeon?},
  author={Connor, Martin J and Dasgupta, Prokar and Ahmed, Hashim U and Raza, Asif},
  journal={Nature Reviews Urology},
  volume={17},
  number={11},
  pages={643--649},
  year={2020},
  publisher={Nature Publishing Group UK London}
}

@inproceedings{oliva2025kidneydepth,
  title={KidneyDepth: A Synthetic Kidney Dataset for Metric Depth Estimation in Ureteroscopy},
  author={Oliva-Maza, Laura and Steidle, Florian and Klodmann, Julian and Strobl, Klaus and Miernik, Arkadiusz and Triebel, Rudolph},
  booktitle={International Conference on Medical Image Computing and Computer-Assisted Intervention},
  pages={331--340},
  year={2025},
  organization={Springer}
}

@article{teufel2024oneslam,
  title={OneSLAM to map them all: a generalized approach to SLAM for monocular endoscopic imaging based on tracking any point},
  author={Teufel, Timo and Shu, Hongchao and Soberanis-Mukul, Roger D and Mangulabnan, Jan Emily and Sahu, Manish and Vedula, S Swaroop and Ishii, Masaru and Hager, Gregory and Taylor, Russell H and Unberath, Mathias},
  journal={International Journal of Computer Assisted Radiology and Surgery},
  volume={19},
  number={7},
  pages={1259--1266},
  year={2024},
  publisher={Springer}
}

@inproceedings{liu2022sage,
  title={Sage: slam with appearance and geometry prior for endoscopy},
  author={Liu, Xingtong and Li, Zhaoshuo and Ishii, Masaru and Hager, Gregory D and Taylor, Russell H and Unberath, Mathias},
  booktitle={2022 International conference on robotics and automation (ICRA)},
  pages={5587--5593},
  year={2022},
  organization={IEEE}
}

@inproceedings{tateno2015real,
  title={Real-time and scalable incremental segmentation on dense slam},
  author={Tateno, Keisuke and Tombari, Federico and Navab, Nassir},
  booktitle={2015 IEEE/RSJ International Conference on Intelligent Robots and Systems (IROS)},
  pages={4465--4472},
  year={2015},
  organization={IEEE}
}

@article{liu2021rds,
  title={RDS-SLAM: Real-time dynamic SLAM using semantic segmentation methods},
  author={Liu, Yubao and Miura, Jun},
  journal={Ieee Access},
  volume={9},
  pages={23772--23785},
  year={2021},
  publisher={IEEE}
}

@article{wu2022semantic,
  title={Semantic SLAM based on deep learning in endocavity environment},
  author={Wu, Haibin and Zhao, Jianbo and Xu, Kaiyang and Zhang, Yan and Xu, Ruotong and Wang, Aili and Iwahori, Yuji},
  journal={Symmetry},
  volume={14},
  number={3},
  pages={614},
  year={2022},
  publisher={MDPI}
}

@inproceedings{lin2023semantic,
  title={Semantic-super: a semantic-aware surgical perception framework for endoscopic tissue identification, reconstruction, and tracking},
  author={Lin, Shan and Miao, Albert J and Lu, Jingpei and Yu, Shunkai and Chiu, Zih-Yun and Richter, Florian and Yip, Michael C},
  booktitle={2023 IEEE International Conference on Robotics and Automation (ICRA)},
  pages={4739--4746},
  year={2023},
  organization={IEEE}
}

@article{umeyama1991least,
  title={Least-squares estimation of transformation parameters between two point patterns},
  author={Umeyama, Shinji},
  journal={IEEE Transactions on Pattern Analysis \& Machine Intelligence},
  volume={13},
  number={04},
  pages={376--380},
  year={1991},
  publisher={IEEE Computer Society}
}

@article{tyszkiewicz2020disk,
  title={DISK: Learning local features with policy gradient},
  author={Tyszkiewicz, Micha{\l} and Fua, Pascal and Trulls, Eduard},
  journal={Advances in Neural Information Processing Systems},
  volume={33},
  pages={14254--14265},
  year={2020}
}

@inproceedings{lindenberger2023lightglue,
  title={Lightglue: Local feature matching at light speed},
  author={Lindenberger, Philipp and Sarlin, Paul-Edouard and Pollefeys, Marc},
  booktitle={Proceedings of the IEEE/CVF International Conference on Computer Vision},
  pages={17627--17638},
  year={2023}
}

@inproceedings{sarlin2019coarse,
  title     = {From Coarse to Fine: Robust Hierarchical Localization at Large Scale},
  author    = {Paul-Edouard Sarlin and
               Cesar Cadena and
               Roland Siegwart and
               Marcin Dymczyk},
  booktitle = {CVPR},
  year      = {2019}
}

@inproceedings{lu2025kidney,
  title={Kidney endoscopy video to preoperative CT alignment for depth estimation},
  author={Lu, Daiwei and Li, Hao and Pierre, Clifford and Kavoussi, Nicholas and Oguz, Ipek},
  booktitle={Medical Imaging 2025: Image-Guided Procedures, Robotic Interventions, and Modeling},
  volume={13408},
  pages={106--112},
  year={2025},
  organization={SPIE}
}

@article{li2025monocular,
  title={Monocular absolute depth estimation from endoscopy via domain-invariant feature learning and latent consistency},
  author={Li, Hao and Lu, Daiwei and d'Almeida, Jesse and Isik, Dilara and Aghdam, Ehsan Khodapanah and DiSanto, Nick and Acar, Ayberk and Sharma, Susheela and Wu, Jie Ying and Webster III, Robert J and others},
  journal={arXiv preprint arXiv:2511.02247},
  year={2025}
}

@article{li2024deep,
  title={Deep Learning--based Unsupervised Domain Adaptation via a Unified Model for Prostate Lesion Detection Using Multisite Biparametric MRI Datasets},
  author={Li, Hao and Liu, Han and von Busch, Heinrich and Grimm, Robert and Huisman, Henkjan and Tong, Angela and Winkel, David and Penzkofer, Tobias and Shabunin, Ivan and Choi, Moon Hyung and others},
  journal={Radiology: Artificial Intelligence},
  volume={6},
  number={5},
  pages={e230521},
  year={2024},
  publisher={Radiological Society of North America}
}

@article{yang2024depth,
  title={Depth anything v2},
  author={Yang, Lihe and Kang, Bingyi and Huang, Zilong and Zhao, Zhen and Xu, Xiaogang and Feng, Jiashi and Zhao, Hengshuang},
  journal={Advances in Neural Information Processing Systems},
  volume={37},
  pages={21875--21911},
  year={2024}
}

@inproceedings{ronneberger2015u,
  title={U-net: Convolutional networks for biomedical image segmentation},
  author={Ronneberger, Olaf and Fischer, Philipp and Brox, Thomas},
  booktitle={International Conference on Medical image computing and computer-assisted intervention},
  pages={234--241},
  year={2015},
  organization={Springer}
}

@inproceedings{jane2026segmentation,
  title={Efficient Labeling: Is Prompt-Based Annotation a Viable Alternative
to Manual Labeling in Endoscopy Video Segmentation?},
  author = {Kanyifeechukwu J. Oguine and Hao Li and Dilara Isik and Ayberk Acar and Jesse d’Almeida and Susheela Sharma and Lauren Branscombe and Josh Petrin and Jie Ying Wu and Robert J. Webster III and Ipek Oguz},
  booktitle={Medical Imaging 2026: Image Processing},
  volume={in press},
  year={2026},
  organization={SPIE}
}

@article{saba2022design,
  title={Design and validation of a non-biohazardous simulation model for holmium laser enucleation of the prostate (HoLEP)},
  author={Saba, Patrick and Shepard, Lauren and Gopal, Narang and Setia, Shaan and Jain, Rajat and Quarrier, Scott and Miller, Nicole and Krambeck, Amy and Humphreys, Mitchel and Ghazi, Ahmed},
  journal={Urology Video Journal},
  volume={16},
  pages={100184},
  year={2022},
  publisher={Elsevier}
}

@article{park1994robot,
  title={Robot sensor calibration: solving AX= XB on the Euclidean group},
  author={Park, Frank C and Martin, Bryan J},
  journal={IEEE Transactions on Robotics and Automation},
  volume={10},
  number={5},
  pages={717--721},
  year={1994},
  publisher={IEEE}
}

\end{document}